\documentclass[runningheads,a4paper]{cmmr2023}
\usepackage[giveninits=true]{biblatex}
\usepackage{times}
\usepackage{amssymb}
\usepackage{amsmath}
\usepackage{graphicx}
\usepackage{subcaption}
\usepackage{booktabs}
\usepackage{siunitx}
\usepackage{bm}

\DeclareFieldFormat{url}{}
\DeclareFieldFormat{doi}{}
\DeclareFieldFormat{issn}{}
\DeclareFieldFormat{isbn}{}
\DeclareFieldFormat{urldate}{}

\newcommand{\keywords}[1]{\par\addvspace\baselineskip
\noindent\keywordname\enspace\ignorespaces#1}

\begin{document}

\mainmatter  

\title{Combining Vision and EMG-Based Hand Tracking for Extended Reality Musical Instruments}
\titlerunning{Multimodal Hand Tracking for Extended Reality Musical Instruments}
\author{Max Graf \inst{1}\and Mathieu Barthet \inst{1} }
%
\authorrunning{Max Graf and Mathieu Barthet}
\institute{Centre for Digital Music, Queen Mary University of London \\ \email{\{max.graf, m.barthet\}@qmul.ac.uk}}

\maketitle

\begin{abstract}
Hand tracking is a critical component of natural user interactions in extended reality (XR) environments, including extended reality musical instruments (XRMIs). However, self-occlusion remains a significant challenge for vision-based hand tracking systems, leading to inaccurate results and degraded user experiences. In this paper, we propose a multimodal hand tracking system that combines vision-based hand tracking with surface electromyography (sEMG) data for finger joint angle estimation. We validate the effectiveness of our system through a series of hand pose tasks designed to cover a wide range of gestures, including those prone to self-occlusion. By comparing the performance of our multimodal system to a baseline vision-based tracking method, we demonstrate that our multimodal approach significantly improves tracking accuracy for several finger joints prone to self-occlusion. These findings suggest that our system has the potential to enhance XR experiences by providing more accurate and robust hand tracking, even in the presence of self-occlusion.
\end{abstract}

\section{Introduction}
Extended reality (XR) is an umbrella term encompassing virtual, augmented and mixed reality (VR/AR/MR). In recent years, the increased popularity of XR technology has seen the establishment of extended reality musical instruments (\textit{XRMIs}) as a research field \cite{serafinVirtualRealityMusical2016}.
Milgram et al. described the reality-virtuality continuum \cite{milgramAugmentedRealityClass1994}, along which digital applications can be placed. It stretches from real-world environments to fully virtual environments. Head-mounted XR devices bridge this continuum. They are capable of rendering three-dimensional imagery onto screens, removing the necessity for separate monitors or mobile displays, blending the real and virtual worlds together. 
The rapid development of XR technologies has opened up new possibilities for musical creation, performance, and interaction, with the emergence of various XR-based musical instruments and applications. Many XRMIs follow an embodied interaction paradigm. These instruments offer novel opportunities for artists to experiment with embodied interaction techniques, spatial sound design, and immersive performances, thus expanding the boundaries of traditional music making.
XRMIs fall within the larger category of digital musical instruments (DMIs). \cite{wesselProblemsProspectsIntimate2001} suggest that the control of DMIs can be made intimate (personal and familiar) by using appropriate control metaphors, low latency action-to-sound, and continuous gesture recognition. This study is part of a larger project that aims to support control intimacy in XRMIs.

Based on the current state of XR technology and prior work in XRMIs \cite{bilbowDevelopingMultisensoryAugmented2021, grafMixedRealityMusical2022}, we highlight that gesture sensing errors on XR devices are a bottlenek for intimate musical control.
Head-mounted XR devices (HMDs) rely on a set of sensors to record data and provide embodied control interfaces for users, e.g., head-tracking, hand tracking, and body pose detection. The transduction of these real-world sensor data to digital representations depends on computational methods.
In this work we focus on the problem of hand tracking, more specifically, accurate tracking of finger joints.
Hand-tracking algorithms often use visual information from camera sensors in conjunction with machine learning techniques, for example, in the Oculus Quest 2 device \cite{hanMEgATrackMonochromeEgocentric2020}. The accuracy of vision-based hand-tracking algorithms may be high \cite{weichertAnalysisAccuracyRobustness2013}, but current recognition rates do not reach 100\%. Self-occlusion - the occlusion of finger joints by other parts of the hand - as well as challenging lighting situations lead to failure cases in vision-based tracking systems.
Such error cases may produce instances of jitter, tracking loss, or glitches in the virtual representation of the hands, which can have detrimental effects on the usability and user experience in XRMIs, as shown in a previous study \cite{grafMixedRealityMusical2022}.

This work aims to address such sensing-related issues through the use of surface electromyography (sEMG) sensors. EMG sensors measure the electrical potential produced during muscle contractions in the body. Surface electromyograms can be obtained through electrodes that are positioned on the surface of the skin, above muscle tissue regions.
We present an investigation into the potential of sEMG sensors and deep learning models to enhance hand-tracking accuracy in XRMIs. Our approach combines sEMG data and vision-based tracking methods to address sensing-related issues commonly encountered in XRMI performance. 
Thereby, we aim to improve the tracking accuracy and responsiveness of XR musical instruments, especially in situations where vision-based tracking falls short.

The scope of this paper is limited to the exploration of sEMG and deep learning techniques for hand-tracking in XR musical instruments. While our findings may have broader applications in other areas of XR interaction, the primary focus is on the improvement of XRMI design and user experience.
Through our work, we aim to contribute to the ongoing development of more accurate, intuitive, and expressive extended reality musical instruments.
\section{Background}
The development of XRMIs has attracted growing interest as VR, AR, and MR technologies continue to advance. Early studies in this domain focused on the design and evaluation of virtual interfaces for musical performance and interaction \cite{maki-patolaExperimentsVirtualReality2005, serafinVirtualRealityMusical2016, fillwalkChromaChordVirtualMusical2015, mooreWedgeMusicalInterface2015}. 
Several works investigated user experience \cite{deaconUserExperienceInteractive2017, camciExploringAffordancesVR2020}, interaction techniques \cite{berthaut3DInteractionTechniques2020} and collaborative music making \cite{menLeMoSupportingCollaborative2018, hamiltonGesturebasedCollaborativeVirtual2016}.
More recent studies have explored the creation of novel instruments and control schemes \cite{camciExploringAffordancesVR2020, bilbowDevelopingMultisensoryAugmented2021, Bilbow2022Evaluating, grafMixedRealityMusical2022}. 
While there is no gold standard for XRMI design, many XRMIs rely on hand-tracking to facilitate embodied interaction with the instrument \cite{mooreWedgeMusicalInterface2015,fillwalkChromaChordVirtualMusical2015,Bilbow2022Evaluating,grafMixedRealityMusical2022}. 
Various vision-based tracking methods are employed, including depth-sensing cameras \cite{mooreWedgeMusicalInterface2015,fillwalkChromaChordVirtualMusical2015,Bilbow2022Evaluating}, and machine learning-based approaches \cite{hanMEgATrackMonochromeEgocentric2020}. Despite the progress in hand tracking research, limitations such as occlusion, lighting issues, and computational complexity continue to pose challenges for hand-controlled XRMI applications.\\

Surface electromyography (sEMG) has emerged as a promising alternative to vision-based tracking methods for capturing user input in various applications, including XR. 
Several studies have explored the use of sEMG data and deep learning architectures for hand gesture recognition \cite{liuNeuroPose3DHand2021, liuWRHandWearableArmband2021, avianEstimatingFingerJoint2022, leeExplainableDeepLearning2022}. These works have reported promising results, highlighting the potential of employing sEMG and deep learning models for precise finger movement estimation. However, some of these works depend on complex tracking setups \cite{avianEstimatingFingerJoint2022} or leverage low-resolution tracking data for training \cite{leeExplainableDeepLearning2022}.

A notable limitation of these studies is the lack of shared training data and code, hindering the reproducibility and comparability of the results across different research efforts. Several datasets on the topic of finger joint angle estimation through sEMG data have been published. However, they either require specialised sEMG measuring equipment \cite{jarque-bouCalibratedDatabaseKinematics2019, kaczmarekPutEMGSurfaceElectromyography2019, jiangGestureRecognitionBiometrics2022}, making the reproduction of results an expensive endeavour, or introduce temporal biases into the dataset due to lack of synchronisation during the recording procedure \cite{huFingerMovementRecognition2022}.
The absence of implementation details (\cite{liuNeuroPose3DHand2021, liuWRHandWearableArmband2021, avianEstimatingFingerJoint2022}) makes it difficult for other researchers to build upon these works, potentially slowing down the progress in the field.

The potential benefits of integrating sEMG data and deep learning models with vision-based tracking methods has not been thoroughly investigated in the context of XRMIs. While machine learning has found its way into the NIME community \cite{theojourdanMachineLearningMusical2023}, the use of machine learning approaches to improve XRMI control remains an underexplored area.
This study aims to develop a multimodal hand tracking approach that leverages sEMG data, deep learning models, and vision-based tracking techniques. We share the training data and code\footnote{\url{https://github.com/maxgraf96/sEMG-myo-unity}}, fostering further research and innovation in the domain of sEMG-based neural interfaces.

\section{Deep Learning Model for Finger Joint Angle Estimation}
\label{sec:data_collection}
We have developed a software pipeline for data collection, feature extraction and modelling of sEMG data. We focus on eight finger joints that are prone to self-occlusion: the metacarpophalangeal and proximal interphalangeal joints of the index, middle, ring and pinky fingers. Specifically, we want to model the rotations of the finger bones connected to these joints relative to the hand. 
The thumb is excluded from our investigation. Modelling thumb rotations with sEMG data is a hard problem, since the majority of muscles related to thumb movements are located in the hand, rather than the forearm.
With that in mind, the goal of the model is to estimate the eight finger joint angles from a window of sEMG data.

\subsection{Data Collection}
We collect surface EMG signal measurements using the Thalmic Labs Myo armband\footnote{\url{https://xinreality.com/wiki/Myo}} and vision-based hand tracking data from the Oculus Quest 2 XR headset. Both devices are employed simultaneously to capture the muscle activity and finger joint rotations, respectively. This approach allows users to capture data without the need for external tracking devices.

The Myo armband is a non-invasive wearable device that features eight sEMG sensors. The armband is worn on the forearm, with the sensors evenly distributed around the circumference of the arm, allowing it to capture the activity of the forearm muscles during finger movements. We obtain sEMG data from the Myo armband using the \textit{PyoMyo} Python framework \cite{Walkington_PyoMyo_2021}, extracting rectified and smoothed signals at a sampling frequency of 50Hz.
The Oculus Quest 2 XR headset is equipped with four monochrome cameras that provide a wide field of view, enabling it to capture hand positions and movements. The built-in hand tracking algorithm \cite{hanMEgATrackMonochromeEgocentric2020} processes the camera data and estimates the 3D rotations of the user's hand joints in real time. In our system we sample the hand joint rotations from the XR device at 50 Hz and synchronize them with the sEMG data from the Myo armband.

One researcher recorded hand gestures and movements in a controlled environment, with a focus on gestures relevant to XRMI interaction. This study should be seen as a proof-of-concept for our methodology. Hence, for this study, we focused on data from the right hand only. The gestures included various finger flexions and extensions, as well as combinations of multiple finger movements. They were performed at different speeds, forces, and orientations. We conducted three data collection sessions across three days to account for the natural variability in sEMG readings, ensuring a more robust dataset.
During the data collection process, the XR headset was strategically positioned in diverse locations and orientations to minimise self-occlusion of the hand. Data collection sessions lasted between ten and fifteen minutes, resulting in a substantial amount of synchronized sEMG and hand tracking data.

\subsection{Feature Extraction}
\begin{figure}[t]
  \centering
  \includegraphics[width=\textwidth]{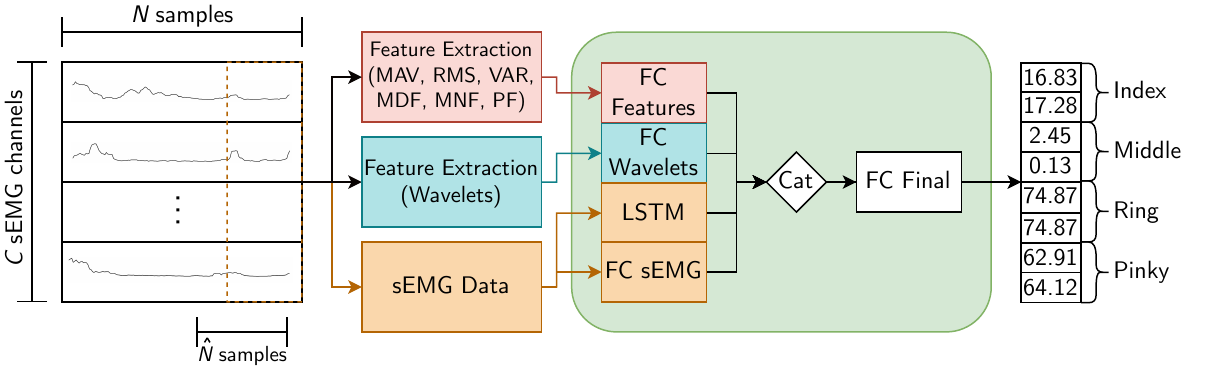}
  \caption{Data preprocessing, feature extraction, and model pipeline}
  \label{fig:pipeline}
\end{figure}

Figure \ref{fig:pipeline} shows the data flow in our pipeline. The selection of features was informed by both the literature and a series of experiments.We use Python to extract both time domain and frequency domain features from the sEMG data. The pipeline takes 2D windows of sEMG samples, with \(N\) number of samples and \(C\) channels. 
We then compute the following features per channel: in the time domain, mean absolute value (MAV), root mean square (RMS), and variance (VAR); in the frequency domain, median frequency bin (MDF), mean frequency bin (MNF), and peak frequency bin (PF). Additionally, wavelet coefficients at the fourth level are extracted to provide further information about the signal's characteristics, as reported in \cite{avianEstimatingFingerJoint2022}. Wavelet analysis provides a multi-resolution representation of the sEMG signal, capturing both the time and frequency characteristics of the data. 
This comprehensive feature representation aims to capture essential information from the sEMG signals both locally and globally, enabling holistic representation of the data.

\subsection{Model Architecture}
The model architecture is a combination of a Long Short-Term Memory (LSTM) network and multiple separate sets of fully connected (FC) layers, aiming to capture both general trends in the sEMG signal data provided by the features and high-frequency characteristics of the signal. 
Formally, our model learns a mapping 
\begin{equation}
F: \mathbb{R}^{N \times C} \rightarrow \mathbb{R}^M
\end{equation}
where \(N\) denotes the number of sEMG samples, \(C\) represents the number of sEMG channels and \(M\) denotes the number of predicted joint angle values at every time step.
More accurately, we learn a mapping
\begin{equation}
      F(\mathbf{s}) = \phi_{\text{final}} \Bigl(\phi_{\text{lstm}}(\mathbf{s}_{\hat{N}}) \oplus \phi_{\text{feat}}(\psi_{\text{time-freq}}(\mathbf{s}_{N})) \oplus \phi_{\text{wav}}(\psi_{\text{wavelet}}(\mathbf{s}_{N})) \oplus \phi_{\text{filt}}(\mathbf{s}_{\hat{N}}) \Bigr)
\end{equation}
where
\begin{itemize}
    \item \(F(\mathbf{s})\) represents the mapping function that takes \(N \times C\) sEMG samples (\(\mathbf{s}\)) and outputs \(M\) finger joint angles.
    \item \(\mathbf{s}_{N}\) denotes all \(N \times C\) sEMG samples.
    \item \(\mathbf{s}_{\hat{N}}\) denotes the last \(\hat{N} \times C\) sEMG samples in the data point.
    \item \(\phi_{\text{lstm}}\) represent the LSTM layers, and \(\phi_{\text{feat}}\), \(\phi_{\text{wav}}\) and \(\phi_{\text{filt}}\) represent the fully connected layers processing time/frequency domain features, wavelet features and filtered EMG data respectively.
    \item \(\phi_{\text{final}}\) denotes the final set of fully connected layers.
    \item \(\psi_{\text{time-freq}}\) and \(\psi_{\text{wavelet}}\) denote the feature extraction functions for time-frequency domain features and wavelet features, respectively.
    \item \(\oplus\) represents the tensor concatenation operation.
\end{itemize}

The LSTM network, a type of recurrent neural network, operates on the subset \(\mathbf{s}_{\hat{N}}\), which contains the last \(\hat{N}\) samples of the EMG data, capturing the temporal dependencies within the most recent portion of the sEMG signal. LSTMs can effectively learn medium-to-long-range dependencies in time-series data and retain information across multiple time steps. 
Additionally, \(\mathbf{s}_{\hat{N}}\) is fed into a separate fully connected layer, which was empirically found to improve the model's performance.
The time and frequency domain features are computed over all \(N\) samples and processed by a set of fully connected layers. These layers are designed to extract higher-level representations from the sEMG features, capturing general patterns and trends in the data. 
The wavelet features are fed into a separate set of fully connected layers, allowing the model to learn distinct patterns associated with the wavelet coefficients. This additional information can help the model to better discriminate between different types of hand movements and gestures.

The outputs of the LSTM layers and the three sets of fully connected layers (time-frequency domain features, wavelet features, and \({\hat{N}}\) sEMG samples) are concatenated and passed to a final set of fully connected layers. This combination of network components aims to capture a comprehensive representation of the sEMG signal, taking into account both general trends and high-frequency changes. The final output of the model is an estimation of the eight finger joint angles described above.

\subsection{Model Implementation and Training} 
We selected \(N=150\), which gives a sampling window size of three seconds. Related works use window sizes of five seconds \cite{liuNeuroPose3DHand2021, liuWRHandWearableArmband2021}. During our experiments with the model architecture, we found that a smaller window size of 150 samples did not reduce the quality of the predicted finger joint angles, while yielding a performance gain in the data processing pipeline. $ \hat{N}=50 $ was selected as a trade-off between LSTM accuracy and performance. Higher values for $\hat{N}$ produced slightly better results, but incurred a performance degradation, slowing down the model at inference time.

The collected data comprises approximately 80000 sEMG samples with corresponding finger joint angle measurements. 
The sEMG samples were separated into training and validation sets using a 90/10 ratio.
Our deep learning model is built using the PyTorch framework.
The model was trained on a single NVidia RTX 2080 Ti GPU for approximately 500000 steps with a batch size of 256. The mean squared error function was employed as a loss metric. We applied an exponentially decreasing learning rate, starting at 0.003 and reducing to 0.0003 over the first 10000 steps.
During training, we tracked the mean average difference between the predicted joint angles and the ground truth angles for both training and validation sets.
Our criterion for stopping the training procedure was the moment of obtaining a mean joint angle difference value of less than 1° across the validation set. 

\section{Multimodal XR Hand Tracking with sEMG and Vision-Based Tracking}
In this section, we present our approach to multimodal XR hand tracking by combining sEMG and vision-based tracking techniques. 
In our system, the vision-based tracking data provides information about the overall hand position and orientation in 3D space, while the sEMG-based model produces granular information about individual finger joint rotations. 
The sEMG data are continuously sampled, preprocessed, and passed to the trained deep learning model to estimate the eight finger joint angles. 
The hand position and orientation data are combined with the estimated finger joint angles to generate a complete hand pose representation.

The deep learning model is optimized for real-time performance, ensuring that the sEMG data can be processed with minimal latency. 
Our system operates at 50Hz, which incurs a latency of 20ms.
For this work, data processing and model inference took place on a consumer notebook. The notebook concurrently runs a python server, responsible for sEMG data aggregation, preprocessing and model inference, and a 3D XR environment on the Unity platform. At every time step, the estimated finger joint angles are transferred from Python to Unity through the low-latency ZeroMQ framework\footnote{\url{https://zeromq.org/}}. 
A video demonstration of our system is available online\footnote{\url{https://www.youtube.com/watch?v=ivl2g2t2oaI}}. It shows a side-by-side comparison of the vision-based tracking system and our multimodal approach.

\subsection{Evaluation}
To validate the effectiveness of our multimodal hand tracking system, we conducted an experimental evaluation, which compared the multimodal tracking to the baseline vision-based hand tracking system provided by the XR device. We simultaneously collected tracking data from the vision-based tracking system and the multimodal tracking system.
A Leap Motion sensor was used to acquire ground truth labels for finger joint angles. The Leap Motion is a high-precision vision-based hand tracking device that captures finger joint angles and hand position in 3D space. The ground truth labels were then compared to the results from both the vision-based and the multimodal tracking system.

\begin{figure}[t]
  \centering
  \includegraphics[width=\textwidth]{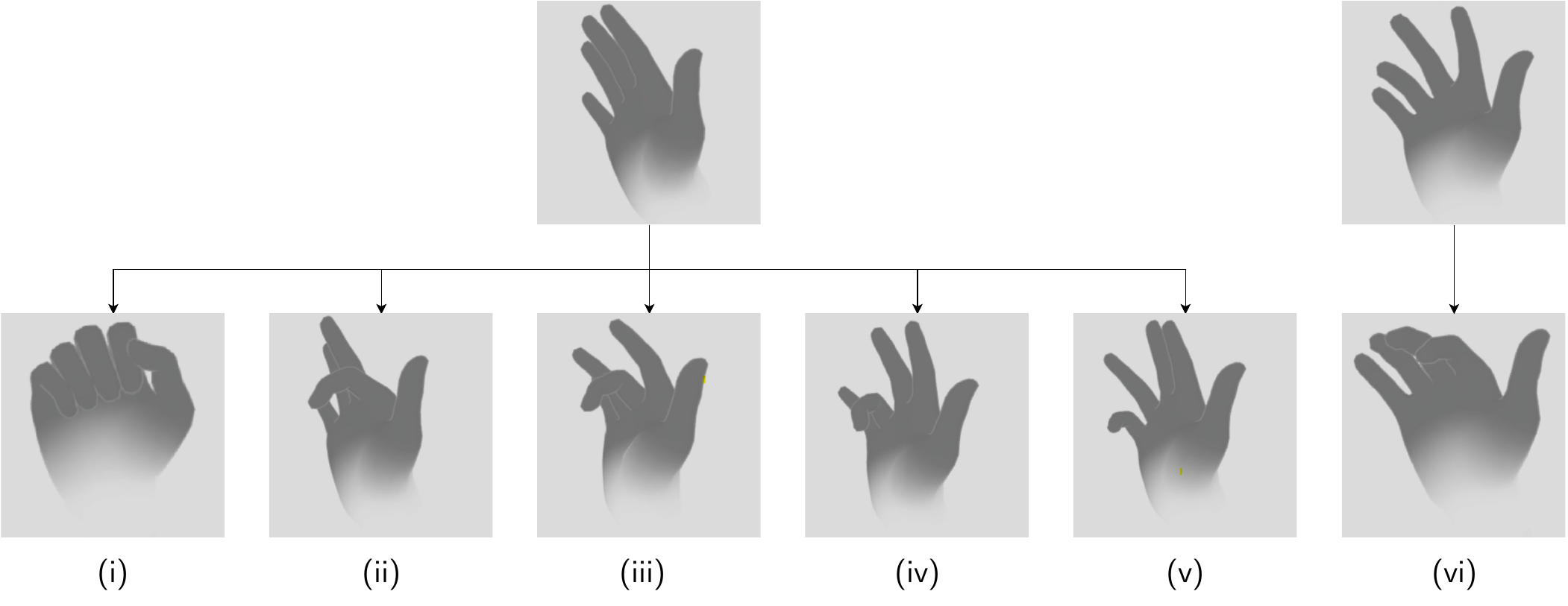}
  \caption{Finger movements in the six hand pose tasks}
  \label{fig:handstask}
\end{figure}

The experimental setup included a series of hand pose tasks, designed to cover a wide range of hand movements, including gestures prone to occlusion. The tasks were selected with regard to their utility in playing a keyboard-inspired XRMI. The tasks were performed while wearing the Oculus Quest 2 headset and the Myo armband. The Leap Motion sensor was placed on a table to record ground truth data. We recorded tasks under two conditions: 1) Full view of the hand - here, the XR headset was positioned at a 50cm distance, 45° above the hand to ensure optimal visual tracking conditions. 2) Self-occlusion of the hand - in this condition, the distance was kept identical, but the angle of the XR headset was lowered, such that the back of the hand occluded the fingers. 
Figure \ref{fig:handstask} illustrates the six hand pose tasks devised: (i) extending all fingers and making a fist; individual flexion and extension of the (ii) index, (iii) middle, (iv) ring and (v) pinky fingers; (vi) sequential flexion and extension of pinky, ring, middle and index fingers (similar to the gesture of drumming on a table while waiting for something). Each task involves the execution of the gesture at three different speeds: slow, over approximately two seconds, moderate (one second), and fast (half a second). To account for variability in the sEMG measurements, all tasks were executed three times, over two days, under identical lighting conditions.

To measure the degree of finger occlusion, we integrated a ray-casting system with the XR application. We cast rays from the XR headset's 3D position to the eight finger bones whose rotations we measured every time a sample was taken. Rays intersecting other parts of the hand, e.g., the back of the hand, were used to mark the respective finger bones as occluded. This allowed us to quantify the level of occlusion per finger for every recording.

\subsection{Analysis Methods \& Results}

To evaluate the performance of the vision-based and multimodal tracking systems across every task, we obtained matrices of difference values between the estimated joint angles and the ground truth angles for both systems at every time step. The matrices were aggregated across the three sessions. We assessed the normality of the difference matrices using the Shapiro-Wilk test for each of the six hand pose tasks.
We then applied the Wilcoxon signed-rank test to see whether there was a significant difference between the results produced by the vision-based and multimodal tracking systems.

\begin{figure}[t]
  \centering
  \includegraphics[width=\textwidth]{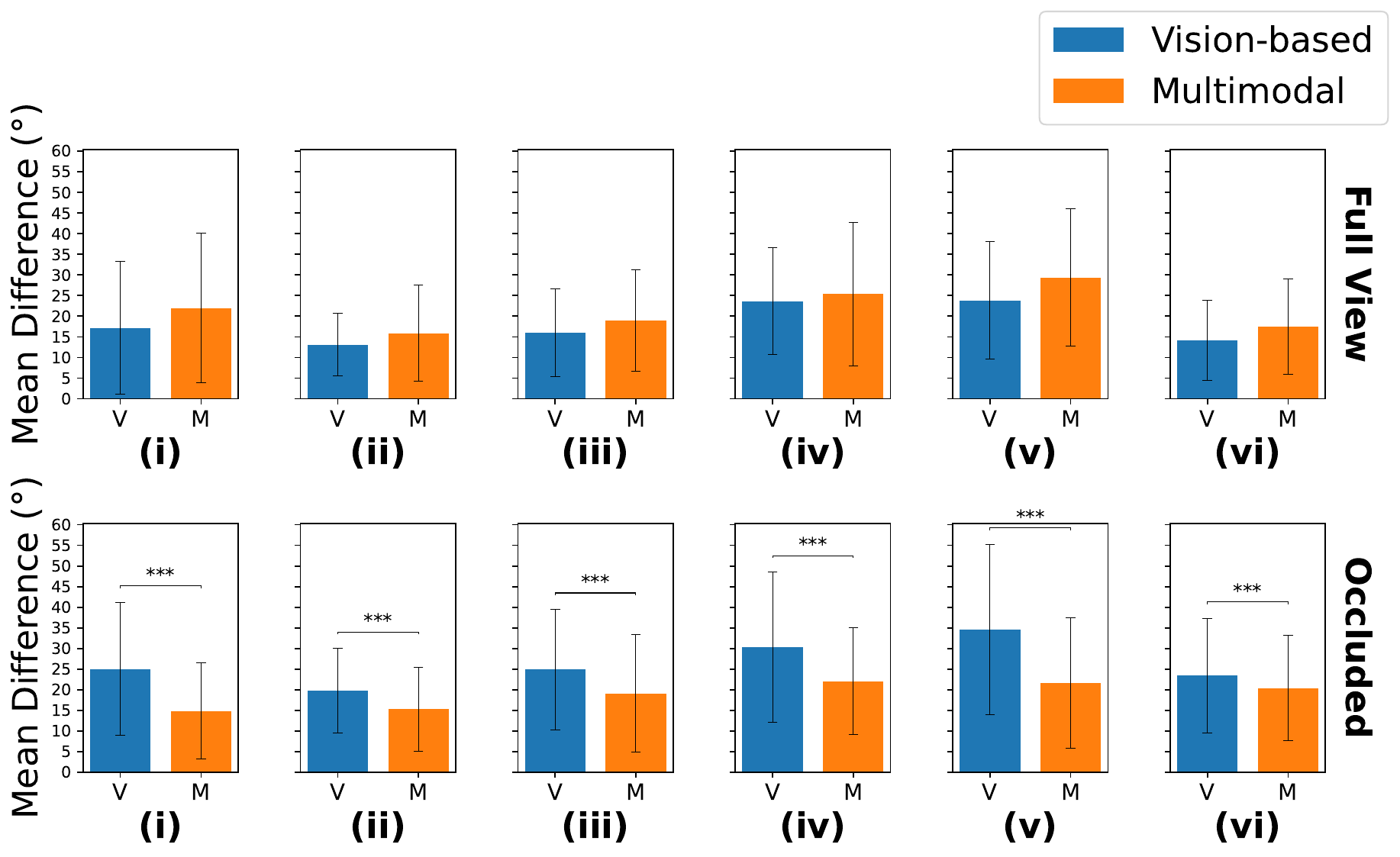}
  \caption{Average deviation in degrees between the finger joint angles generated by the vision-based (V) and multimodal (M) tracking systems and the ground truth data for each task. Error bars show standard deviation; three asterisks indicate a significant difference between the V and M values ($p < .001$)}
  \label{fig:valtestresults}
\end{figure}

Across tasks, the results of the Shapiro-Wilk test showed p-values \(< .001\), indicating that the data in the difference value matrices did not fit a normal distribution. Therefore, we proceeded with the non-parametric Wilcoxon signed-rank test for further analysis.
Figure \ref{fig:valtestresults} shows the results obtained from the mean joint angle differences across all finger joints per task, for both occlusion conditions (full view and occluded). Under the occluded condition, our model produces significantly lower deviations from the ground truth data across all tasks, compared to the vision-based tracking system. On average, it improves the finger joint angle tracking accuracy by five to 15 degrees across all fingers.

Table \ref{table:wilcoxon_pvalues} shows the results obtained from the Wilcoxon signed-rank tests, aggregated across all eight tracked joints per task for both conditions. The p-values indicate significant differences between the difference value matrices. Additionally, the table lists the average finger occlusion results obtained through the raycasting occlusion measure described above. The occlusion results describe the mean portion per task, in which the fingers were occluded by another part of the hand.

\begin{table}[t]
  \centering
  \caption{P-values and average occlusion measurement results across tasks under both conditions}
  \begin{tabular}{l|c|c|c|c}
    \toprule
    & \multicolumn{2}{c|}{\textbf{Full view}} & \multicolumn{2}{c}{\textbf{Occluded}} \\
    \cmidrule{2-5}
    \textbf{Tasks} & P-value & Occlusion (\%) & P-value & Occlusion (\%) \\
    \midrule
    (i)   & 1.0   & 16.78 & \textless .001 & 93.00 \\
    (ii)  & 1.0   & 2.65  & \textless .001 & 70.58 \\
    (iii) & 1.0 & 15.25 & \textless .001 & 63.11 \\
    (iv)  & 1.0  & 16.78 & \textless .001 & 53.02 \\
    (v)   & 1.0   & 27.94 & \textless .001 & 78.82 \\
    (vi)  & 1.0   & 10.36 & \textless .001 & 59.48 \\
    \bottomrule
  \end{tabular}
  \label{table:wilcoxon_pvalues}
\end{table}

\section{Discussion}
The results of the evaluation showed that the multimodal hand tracking system outperformed the pure vision-based hand tracking system across all tasks under the occluded condition. These findings support our hypothesis that the integration of sEMG-based finger joint angle estimation can help overcome occlusion-related limitations in vision-based hand tracking, resulting in more accurate and reliable XRMI interactions.
Under the full view condition, the vision-based hand tracking produced fewer errors in all tasks. This was expected, as the vision-based hand tracking system operates optimally under full view of the hand.

Despite the promising results, our study has several limitations. 
Due to the nature of sEMG data, the tracking performance of the multimodal approach is unlikely to extend to other users without fine-tuning the deep learning model. Surface EMG signals differ substantially between individuals and can be influenced by factors such as muscle fatigue, electrode placement, and individual anatomical differences. It will be valuable to investigate the system's performance across different users and under varying conditions. The identification of sEMG data representations that allow for generalisation under consideration of these factors without requiring extensive amounts of data is still an ongoing research topic. However, our work allows XR users with access to sEMG devices to train their own models using our pipeline and code. \\
The performance of our multimodal hand tracking system was evaluated using a single type of XR headset and sEMG armband. Future research should explore more complex occlusion scenarios, as well as test the system's performance across different hardware setups and sEMG devices, to better understand the generalisability of our findings.\\
With that in mind, we see numerous avenues for further research. The integration of additional tracking modalities, such as depth sensing or inertial measurement units (IMUs), could further enhance the robustness and accuracy of the multimodal hand tracking system by enabling stronger representations of the underlying data.
A future study will explore the impact of our multimodal hand tracking system on usability, user experience and task performance in XRMI interactions, and provide insights into the practical implications of our findings. By conducting user studies with tasks that require precise hand movements and are susceptible to occlusion, the benefits of our system for real-world applications could be better understood.\\
Our study provides evidence that the combination of vision-based tracking and sEMG-based finger joint angle estimation can effectively address occlusion issues in hand tracking for XRMI interactions. The findings suggest that the multimodal hand tracking system has the potential to enhance user experiences and enable more immersive and natural interactions in virtual environments.

\section{Conclusion}
In this paper, we introduced a multimodal hand tracking system designed to address occlusion issues in XRMI interactions by combining vision-based tracking with sEMG-based finger joint angle estimation. The goal of this study was to demonstrate the potential of our proposed system to improve hand tracking accuracy and robustness, even when the hand is partially occluded.

While our results show promise, the experimental setup was relatively simple, and further research should explore more complex scenarios and investigate the system's performance across different hardware and user conditions. Future work could also integrate additional tracking modalities and machine learning techniques to enhance the robustness and accuracy of the system.

Our multimodal hand tracking system demonstrates the potential to improve XRMI interactions by addressing occlusion issues in vision-based hand tracking. As XR technologies continue to evolve, the integration of complementary tracking modalities, such as sEMG and vision-based tracking, will likely play a crucial role in enhancing user experiences and enabling more immersive and natural interactions in virtual environments.


\printbibliography

\end{document}